\documentclass[journal]{IEEEtran}

\ifCLASSINFOpdf
\else
\fi

\usepackage{graphicx}
\usepackage{lipsum} %\lipsum[1-6]

\usepackage{soul}
\usepackage{xcolor}
\sethlcolor{yellow}
\usepackage[caption=true,font=normalsize]{subfig}
\begin{document}

%\title{Bare Demo of IEEEtran.cls\\ for IEEE Journals}
\title{Machine Learning-Based Path Loss Modeling with Simplified Features}


\author{Jonathan~Ethier~and~Mathieu Ch\^{a}teauvert% <-this % stops a space
\thanks{J. Ethier and M. Chateauvert are with the Communications Research Centre, Kanata, Ontario, Canada,  e-mail: jonathan.ethier@ised-isde.gc.ca}% <-this % stops a space
\thanks{}}

\maketitle

\begin{abstract}
Propagation modeling is a crucial tool for successful wireless deployments and spectrum planning with the demand for high modeling accuracy continuing to grow. Recognizing that detailed knowledge of the physical environment (terrain and clutter) is essential, we propose a novel approach that uses environmental information for predictions. Instead of relying on complex, detail-intensive models, we explore the use of simplified scalar features involving the total obstruction depth along the direct path from transmitter to receiver. Obstacle depth offers a streamlined, yet surprisingly accurate, method for predicting wireless signal propagation, providing a practical solution for efficient and effective wireless network planning.
\end{abstract}


% \fi
\IEEEpeerreviewmaketitle

\section{Introduction}
\IEEEPARstart{P}{ropagation} modeling is used to make predictions of wireless signal power levels in aid of radio deployments including coverage and interference analysis. Path loss modeling is a popular approach to this end, allowing for a generic prediction of loss along a wireless link, thereby providing network planners with the ability to consider a range of physical layer attributes (antenna gains, system technology, and the like). In mid-band communications scenarios with mostly line-of-sight (LOS) conditions, propagation modeling can be covered by models such as Longley-Rice \cite{LR} and even the commonly used free space path loss (FSPL) model. 

However, in non-LOS conditions, these models provide degraded prediction accuracy by failing to account for the signal attenuation and interference caused by electromagnetic interactions with terrain and clutter \cite{HATA}. Various propagation models explicitly account for clutter, including ITU-R P.1812-6 \cite{P1812} (hereafter known as P.1812), which is appropriate for mid-band frequencies. These models require knowledge of the entire path profile between transmit and receive, including the complete spatial variation of the terrain (DTM) and surface (DSM) as separate entities (effectively a 1D problem). Other models exist that use thousands of features in 2D \cite{NewPL_Paper} and 3D \cite{Alice} representations of terrain and clutter and perform point-to-multi-point type predictions.

The question addressed in this paper is as follows: can simple features derived from path profiles be used as the sole input to a predictor of path loss along a wireless link and still provide sufficient accuracy for predicting radio coverage? In particular, we will (1) use machine learning (ML)-based modeling, comparing against traditional approaches and (2) emphasize the use of measurement data for training to provide us with reliable ground truth. Similar work has been done \cite{MinFeatures}, but did not use features that describe the physical environment, which we will use gainfully here.

In Section \ref{sec:dataprep}, we will discuss the data sources and processing techniques that can be used to sculpt data to train our models. Following this, in Section \ref{sec:selection}, we will consider the models used in our analysis, along with feature selection and their possible configurations. Section \ref{sec:results} will outline our training and testing approaches and provide the results of the study, followed by Section \ref{sec:obsdepthmodel} where we dive deeper into the influence of the key model feature of obstacle depth. Finally, Section \ref{sec:conclusion} will provide conclusions and propose future work.

%%%%%%%%%%%%%%%%%%%%%%%%%%%%%%%%%%%%%%%%%%%%%%%%%%%%%%%%%%
\section{Data Preparation}
\label{sec:dataprep}

\subsection{Drive Test Measurement Data}
\begin{figure}[ht]
\centering
\includegraphics[width=0.49\textwidth]{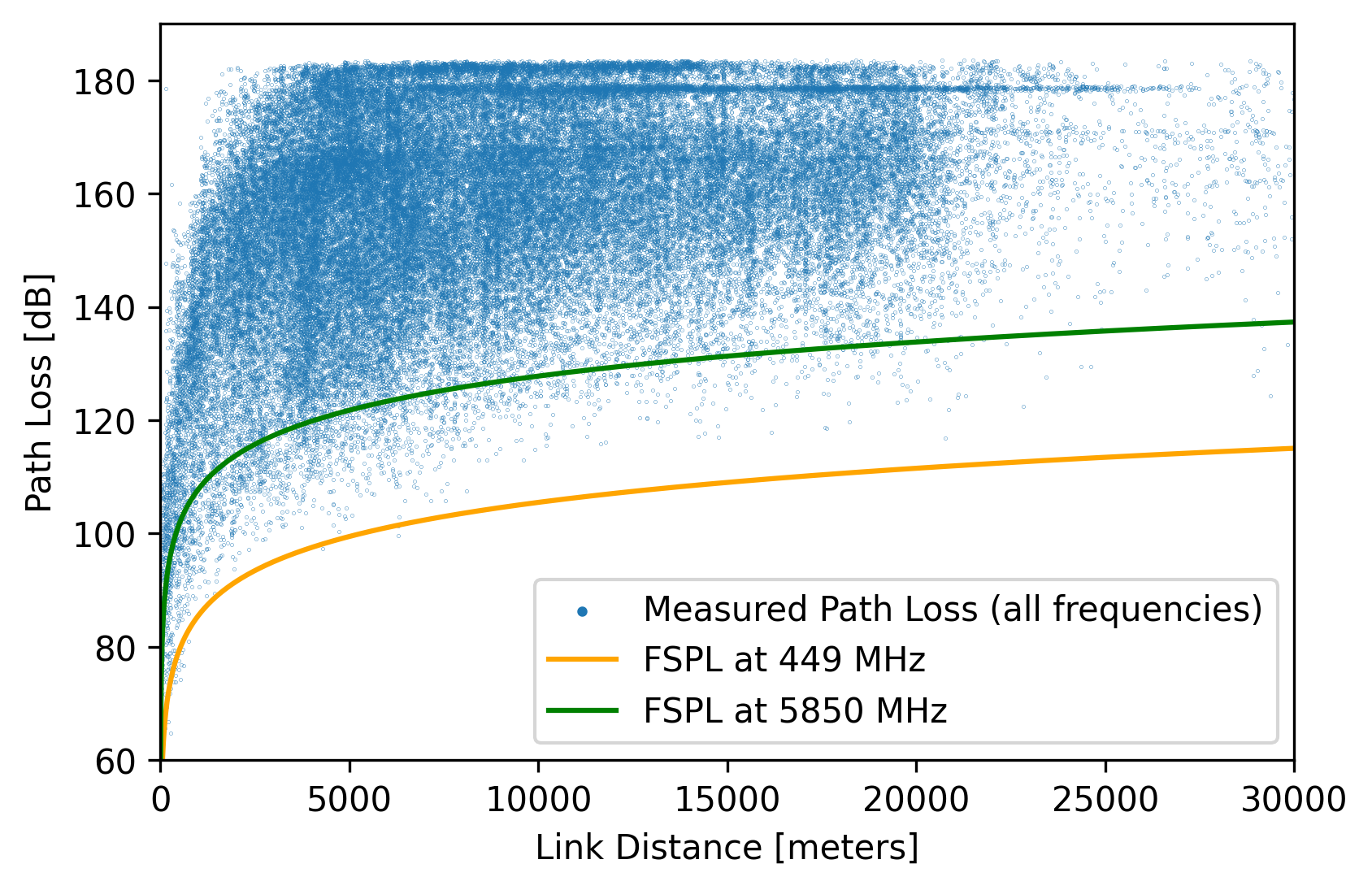}
\caption{Measured Path Loss vs. Distance for the Ofcom \cite{OFCOM_DATA} Data (FSPL shown for the two frequency extremes)}
\label{fig:PL_distance}
\end{figure}

Measurement data is critical when developing new propagation models, namely for model testing. This requirement is exacerbated when training ML-based models, with measurement data taking on double duty: both training and testing. Not only must the data be high quality, but numerous and varied as well. To this end, we use the openly available ITU-R UK Ofcom drive test dataset \cite{OFCOM_DATA}. This drive test was conducted across the United Kingdom from 2015 to 2018, consisting of six frequencies in mid-band (449, 915, 1802, 2695, 3602, and 5850 MHz) at seven geographically and morphologically distinct sites. The dataset consists of 8.2 million measurements, with more than seven million measurements above the measurement noise floor. The distribution of path loss over link distance for the Ofcom data is shown in Fig. \ref{fig:PL_distance}. The reader is encouraged to refer to \cite{OFCOM_DATA} for drive test path details.

The minimum, median, and maximum path loss and link distance in the dataset are 50.64, 160.28, and 183.59 dB, and 18, 8093, and 78613 meters, respectively (only measurements above the noise floor). Transmitter height varies from 17 to 25 meters, while the receiver has a constant height of 1.5 meters. Both transmit and receive use omnidirectional antennas. The majority (98.5\%) of the samples have NLOS conditions. Every sample has a measured path loss in dB, frequency in MHz and the associated latitude, longitude and antenna height above terrain for the receiver and transmitter. A propagation model can be constructed using these parameters but such a model would lack information about the environment. The acquisition of environmental information will be discussed next.

\subsection{Terrain and Surface Databases}
Online databases of DTM and DSM have increased in both detail and coverage in recent years, providing a treasure trove of information for many scientific and engineering disciplines. These databases are made publicly available by governments worldwide, including Canadian \cite{HRDEM}, USA \cite{USGS}, and UK \cite{UK_DTM_DSM} governments. Fortunately, the latter database provides the resources one needs to extract path profiles for six of the seven sites in the Ofcom drive test data (Scar Hill GIS data not being publicly available). The DTM and DSM data are available in geotiff format, making direct path calculations and assessment of terrain and clutter a straightforward task.

\subsection{The SAFE Tool for Path Profile Extraction and Analysis}
The propagation model known as the Signal Attenuation through Foliage Estimator (SAFE) tool \cite{SAFE_PAPER} contains several sub-routines that can be used to process geotiff files to extract the DTM and DSM path profiles, given only the transmit and receive antenna heights and their (lat, lon) coordinates. With path profiles in hand, numerous features can be derived including quantities such as \textit{total obstacle depth} along the direct path. The source code for the SAFE tool is available on GitHub \cite{SAFE_GITHUB} and is used extensively in this paper for path profile and feature extraction.

%\begin{figure}[ht]
%\centering
%\includegraphics[width=0.49\textwidth]{Figures/drivetests.png}
%\caption{Drive Test Locations and Paths from \cite{OFCOM_DATA}}
%\label{fig:drivetests}
%\end{figure}

%%%%%%%%%%%%%%%%%%%%%%%%%%%%%%%%%%%%%%%%%%%%%%%%%%%%%%%%%%

%%%%% PATH PROFILE IMAGE
\begin{figure*}[!t]
\centering
\includegraphics[width=0.9\textwidth]{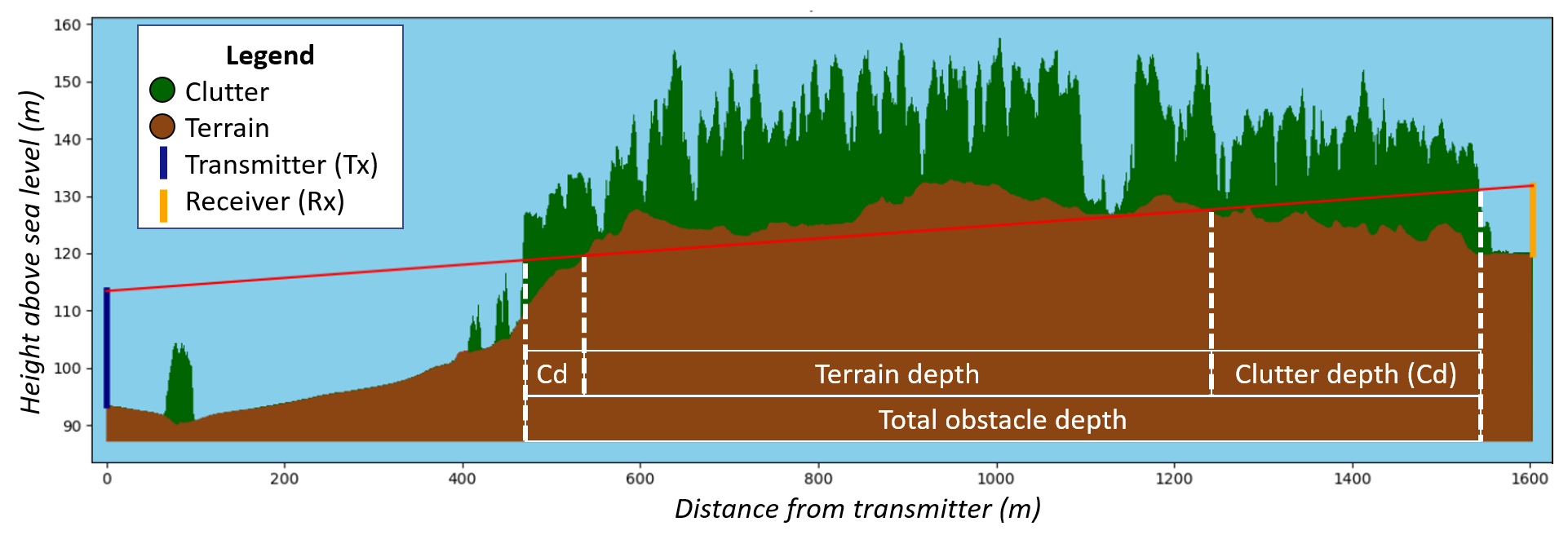}
\caption{Example path profile extracted using the SAFE tool. Total depth through terrain (724 meters) and total depth through clutter (67 meters + 314 meters = 381 meters) are as shown with total obstacle depth as the sum of the two (724 meters + 381 meters = 1105 meters). Any number of non-contiguous blocks can contribute to the total terrain, clutter and obstacle depths.}
\label{fig:pathprofile}
\end{figure*}

\section{Feature and Model Selection}
\label{sec:selection}

%\subsection{Features and their Configurations}

The goal of our feature selection is to use as few features as possible while still sufficiently describing the physical properties of a radio link. The default features in any propagation model are frequency and link distance (distance along the direct path from transmitter to receiver). We consider two additional environmental features: terrain and clutter depth, where depth is calculated as the total obstruction (in meters) in either terrain or clutter along the direct path. Note that the total depth is computed from any number of non-contiguous obstructions, and clutter that appears above obstructing terrain does not contribute to the total clutter depth.

Three configurations of these features will be considered as shown in Table \ref{tab:feature_table}. Note that Obstacle Depth is equal to Terrain Depth + Clutter Depth, i.e. $o = t + c$, and was selected as a simple feature that can be derived from datasets that only contain surface data (DSM) and no separate terrain data (DTM). A visualization of the path profile and feature extraction is shown in Figure \ref{fig:pathprofile}. The three model architectures will be discussed next.

%%% TABLE OF CONFIGS
\begin{table}[ht] % [h] to make the table appear inline instead of the top [!t] sends it to the top
\renewcommand{\arraystretch}{1.5} % Optional: adjusts the row height
\centering
\caption{Feature Configurations}
\begin{tabular}{|p{1.1cm}|p{6.2cm}|} %\begin{tabular}{|p{3cm}|p{4cm}|p{2cm}|}
\hline
\centering{Features} & Feature List  \\ \hline
\centering{2} & Link Distance (d), Frequency (f) \\ \hline
\centering{3} & Dist. (d), Freq. (f), Obstacle Depth (o)  \\ \hline
\centering{4} & Dist. (d), Freq. (f), Terrain Depth (t), Clutter Depth (c)  \\ \hline
\end{tabular}
\label{tab:feature_table}
\end{table}

%\subsection{Modeling Approaches Considered}

\subsubsection{Curve-fit Log Regression (Log-Reg) Model}
The modeling approach we begin with is a traditional curve fit using the scaled logarithms shown in Equations 1-3 for the feature configurations 1-3. Using the $curve\_fit$ method from the Python package $scipy$, we optimize the scalar parameters A-E, depending on the configuration used.
\begin{equation} PL = \mathrm{A}\log_{10}(f) + \mathrm{B}\log_{10}(d) + \mathrm{C} \end{equation} 
\begin{equation} PL = \mathrm{A}\log_{10}(f) + \mathrm{B}\log_{10}(d) + \mathrm{C}\log_{10}(o) + \mathrm{D} \end{equation}
\begin{equation} PL = \mathrm{A}\log_{10}(f) + \mathrm{B}\log_{10}(d) + \mathrm{C}\log_{10}(t) + \mathrm{D}\log_{10}(c) + \mathrm{E} \end{equation}

For brevity's sake, the ideal ML architectures were determined with grid search and the results will be stated here. The model architectures were determined to yield lower prediction errors while not overfitting.

\subsubsection{Boosted Trees (XGBoost) Model}
For boosted trees, the XGBoost Python package was used with the ideal model:
\begin{itemize}
\item
100 regression trees with a depth of 2, with all other parameters left as default. 
\end{itemize}

\subsubsection{Fully-Connected Network (FCN) Model}
Lastly, we consider a fully connected network (FCN) using Tensorflow \& Keras Python packages. The ideal fully connected network has the following hyperparameters (obtained through grid search):
\begin{itemize}
\item One layer of 256 neurons, ReLU activation, a dropout of 20\% after the single layer, ADAM optimization, MSE loss function, batch size of 8,192, an 80/20 random train/validate split and a max of 20 training epochs. 
\end{itemize}

Note that XGBoost and FCNs are both known to perform well with tabular data \cite{Goodfellow-et-al-2016}, which makes them ideal for our scalar feature-based modeling approach.

%%%%% SUMMARY TABLE OF MODEL PERFORMANCE
%\begin{figure*}[!t]
%\centering
%\includegraphics[width=0.95\textwidth]{Figures/example_table2.png}
%\caption{RMS Error Results of City Round Robin}
%\label{fig:tableofresults}
%\end{figure*}

%%%%%%%%%%%%%%%%%%%%%%%%%%%%%%%%%%%%%%%%%%%%%%%%%%%%%%%%%%
\section{Training and Results}
\label{sec:results}

\subsection{Train and Test Approach}
With a large collection of training samples, it is important to carefully select the training and test scenarios to avoid overfitting and data leakage, thus allowing proper assessment of model generalization \cite{bishop2006pattern}. Various test scenarios were considered, including splitting test sites geographically (e.g. training on 80\% of the samples in a city and testing on the remaining non-overlapping 20\%). However, such a scenario allows too much of a drive test to be present in the training data and it would be difficult to judge to what extent the geographic adjacency of measurements influences scores. Alternatively, we take a more stringent route and perform a round-robin of city holdouts, training on five cities and testing on the sixth (repeating this six times). This minimizes both geographic adjacency and data leakage.

\subsection{Test Results}
A summary of the prediction errors for the three modeling techniques, three feature configurations and six city holdouts is shown in Table \ref{tab:test_results_table}. Root mean squared error (RMSE) is used to judge the models as it penalizes large errors more so than mean absolute error (MAE). Additionally, a column listing the RMSE for the FSPL and P.1812-6 models, with the latter quoted from \cite{P1812}, are included as baselines. The P.1812 model performs well on the data, which is to be expected since it was developed using the Ofcom drive test data for model validation. Conversely, an FSPL model performs poorly on this data since the datasets are heavily dominated by NLOS measurements. The coefficients of the Log-Reg London holdout models (Eq. 1-3) are shown in Table \ref{tab:logreg_coeff}. 

\begin{table}[ht]
\renewcommand{\arraystretch}{1.5} \centering
\caption{Log-Reg Model Coefficients (London Holdout)}
\begin{tabular}{|l|l|l|} \hline
No. of Features & Coefficient & Optimized Values \\ \hline
2 features & A, B, C & 30.52, 31.08, -253.24 \\ \hline
3 features & A, B, C, D & 18.32, 31.77, 11.03, -242.34 \\  \hline
4 features & A, B, C, D, E & 20.93, 31.76, 3.96, 8.50, -247.91 \\  \hline
\end{tabular}
\label{tab:logreg_coeff}
\end{table}

For the FCN model, we perform 20 optimization runs for each city/feature combination (18 in total) with random starting weights and train/validation splits. Consequently, the RMSE shown in the table for FCN is the median RMSE among 20 runs, with the largest standard deviation among the six city holdouts and three different feature configurations never exceeding 2\% of the median RMSE. This indicates the modeling is largely insensitive to the starting weights and random train/validate split. This is at least in part due to the very large number of samples that help reduce the variation from run to run. Note that running any single model has a similar maximum RMSE difference of 2\% relative to a 20-model median ensemble, making the ensemble approach not necessarily worth the additional computational costs.

\begin{table*}[!t]
\renewcommand{\arraystretch}{1.5} % Optional: adjusts the row height
\centering
\caption{City Holdout Model Test Results (RMSE in dB)}
\begin{tabular}{|l|c|c|ccc|ccc|ccc|}
\hline
&  &  & \multicolumn{3}{|c|}{Log-Regression} & \multicolumn{3}{|c|}{ Boosted Trees (XGBoost)}  & \multicolumn{3}{|c|}{ Fully-Connected Network (FCN)}  \\
& FSPL & P.1812 & 2 features & 3 features & 4 features & 2 features & 3 features & 4 features & 2 features & 3 features & 4 features \\ \hline
London & 45.7 & 8.8 & 11.65 & 7.44 & 7.59 & 11.04 & 7.75 & 7.60 & 10.66 & 7.44 & 7.58 \\ 
Merthyr & 43.5 & 13.4 & 10.88 & 9.17 & 8.97 & 12.34 & 6.91 & 6.89 & 12.28 & 7.22 & 7.02 \\ 
Nottingham & 40.5 & 12.6 & 11.94 & 7.38 & 7.82 & 12.06 & 7.15 & 6.93 & 12.06 & 7.07 & 6.68 \\ 
Southampton & 47.1 & 9.5 & 11.37 & 7.13 & 6.71 & 10.90 & 6.38 & 6.06 & 10.95 & 6.64 & 6.12 \\ 
Stevenage & 45.7 & 12.3 & 11.46 & 9.13 & 9.24 & 11.17 & 8.63 & 8.49 & 11.14 & 8.81 & 8.63 \\ 
Boston & 31.9 & 11.4 & 17.68 & 7.64 & 8.08 & 17.15 & 8.21 & 7.78 & 17.51 & 7.56 & 7.52 \\  \hline
Median & 44.6 & 11.85 & 12.50 & 7.98 & 8.07 & 12.44 & 7.51 & 7.29 & 12.43 & 7.46 & 7.26 \\ \hline
\end{tabular}
\label{tab:test_results_table}
\end{table*}

The FCN has a lower median RMSE than boosted trees and log regression, though the advantage of the FCN models over XGBoost is not very significant. In all cases, the more features present in the model, the lower the RMSE. For the FCN model, we observed 4.97 and 5.17 dB improvements for three and four features over the simplest two-feature model, respectively. Introducing obstacle depth as a third feature has a very significant impact on reducing RMSE.

Separating obstacle depth into two features (terrain \& clutter) does not provide a significant reduction in error. For the FCN model, we only see a 0.2 dB improvement. This is possibly due to a temporal mismatch between measurements (2015-2018) and geospatial information (2022), which degrades the efficacy of clutter information since it changes over time.
%  \item Despite their simplistic nature, the log regression models managed a respectable performance, lagging behind the ML models with an additional 0.7-0.8 dB RMSE.
%  \item Both ML models had a reduction in RMSE for all six cities with every additional feature added. This was not true for the Log-Reg model, indicating a limitation of the model architecture relative to the ML varieties.

%In terms of model generalization and the round-robin approach to model testing, the London holdout model performs well despite London being the only major city on the six-city list.

The majority of measurements reside in high-clutter urban and suburban environments (98.5\% of measurements are NLOS), with a small number of rural measurements residing at the periphery of the drive test areas. This allows models to generalize well between city holdouts due to the similarity in clutter types and hence the similar RMS errors. One exception is the Stevenage holdout city model which has a higher RMSE relative to the other holdouts. This is likely due to the city having significantly newer construction \cite{OFCOM_DATA} than other drive test sites, thus being more difficult to generalize using training data from cities with older infrastructure. Scar Hill is the only majority rural drive test in the Ofcom measurements, and detailed DTM/DSM data for this region is not currently available in open source and was therefore not included in this study.

\begin{figure}[ht]
\centering
\includegraphics[width=0.49\textwidth]{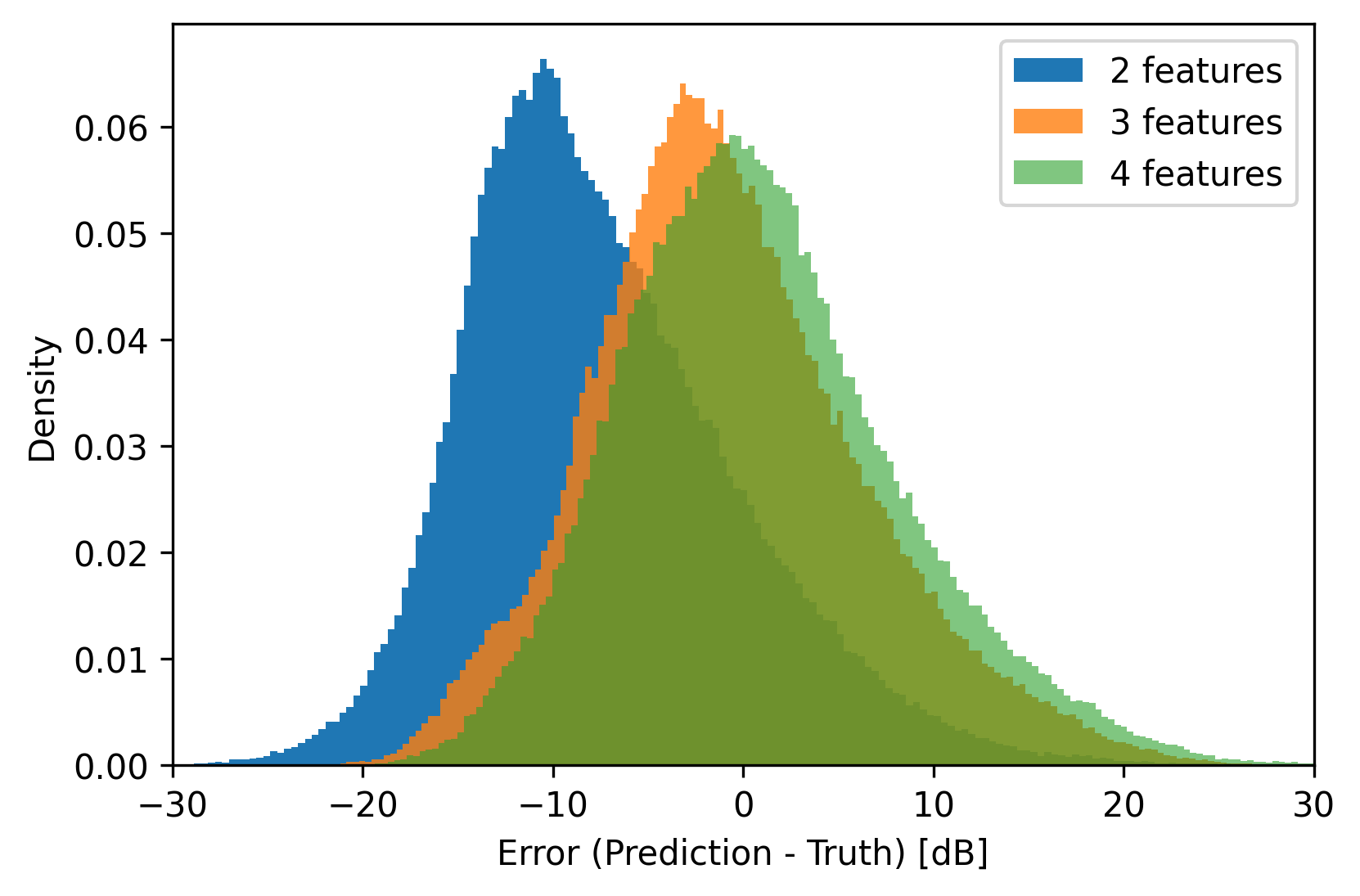}
\caption{Histogram of Prediction Errors (London Holdout, FCN)}
\label{fig:hist_errors}
\end{figure}

A histogram plot of the prediction errors for the fully connected network models (London holdout) is shown in Figure \ref{fig:hist_errors} for two, three and four feature inputs (with -7.70, -0.97 and +0.74 dB median error, respectively). Without environmental information, the predictions under-predict path loss, but as one includes additional features describing the environment, the median error is closer to zero. Similar results can be observed for the other five city holdout test scenarios.

%%%%%%%%%%%%% PLACEHOLDER IMAGE

The RMSE obtained for the path loss model in \cite{NewPL_Paper} is in the range of 3-5 dB whereas our models are in the range of 6-8 dB. In the former, the authors used high-resolution imagery and detailed path profiles (equivalent to many thousands of features), whereas our work relied on 3-4 scalar feature inputs. This comparison emphasizes the engineering balance of increasing model accuracy at the cost of increasing model complexity. Our work shows that on the low end of model complexity, accurate modeling is still feasible.

%Lastly, the two ML models were computationally benchmarked using an A100 GPU, an AMD Epyc 7251 and an SSD drive. Including the extraction of the GIS features (clutter and terrain depths) and ML inference, the prediction speed is approximately 8000 predictions/s for the XGBoost model and 6000 predictions/s for the ML-FCN Tensorflow model. Note that the XGBoost model is slightly faster, albeit with slightly higher RMSE as compared to the ML-FCN Tensorflow model.

\section{Analysis of Obstacle Depth Modeling}
\label{sec:obsdepthmodel}

Given the significant improvement offered by the \textit{total obstacle depth} feature (5 dB reduction in RMSE), it is worthwhile exploring how this feature is accounted for in the ML models, and the FCN model in particular. We can compute the obstacle loss by subtracting the FSPL from the PL predicted by the ML model, as shown in Equation 4:

\begin{equation} 
Obstacle\hspace{1mm}Loss = PL_{ML} - FSPL
\end{equation} 

\begin{figure}[ht]
\centering
\includegraphics[width=0.49\textwidth]{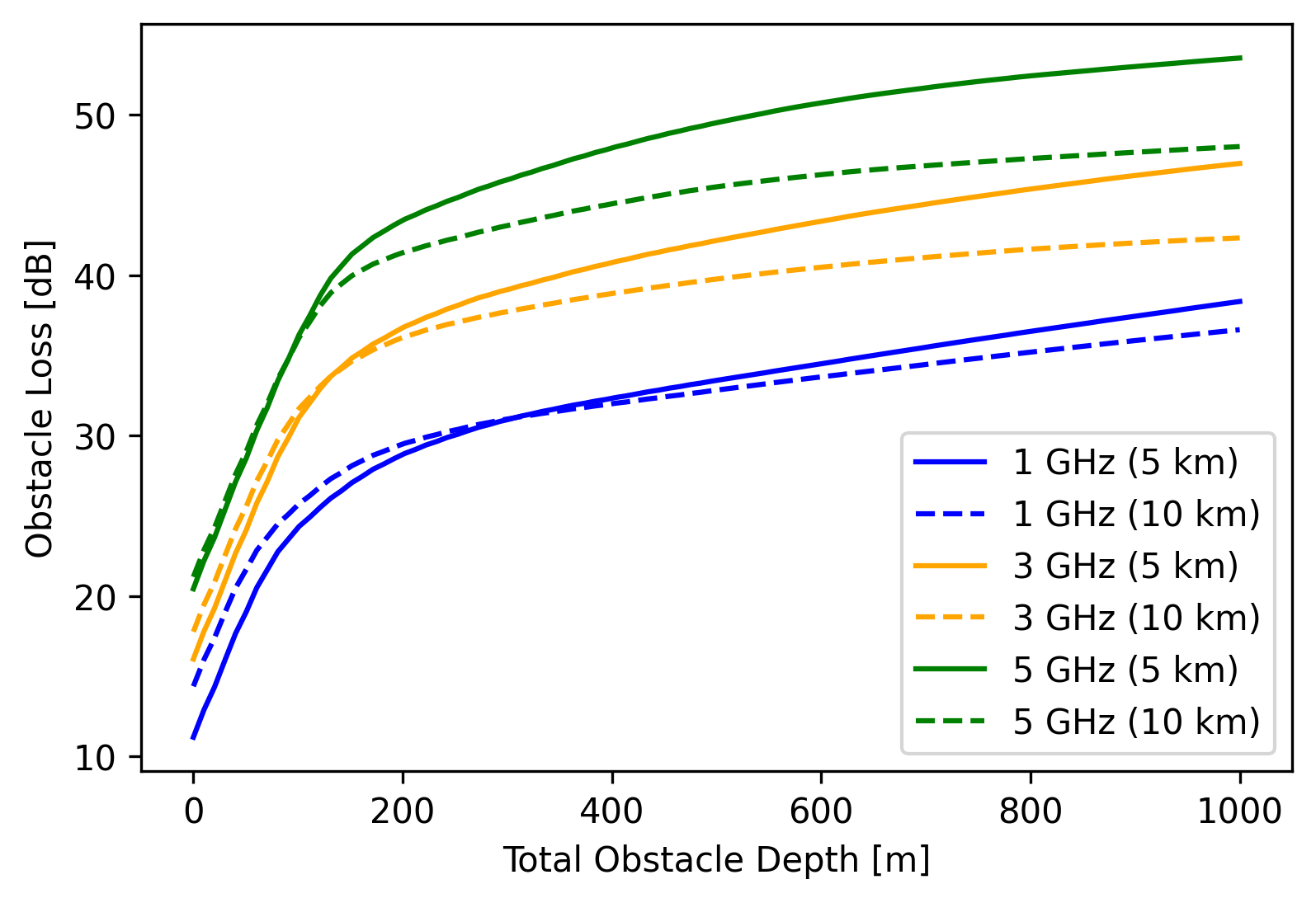}
\caption{Predicted Obstacle Loss for Various Frequencies and Link Distances (London Holdout FCN-ML model)}
\label{fig:obstacle_loss}
\end{figure}

In Figure \ref{fig:obstacle_loss}, we plot the obstacle loss as a function of total obstacle depth, for a few pairs of frequency \& distance. This effectively turns the propagation model into an obstacle loss model. The loss curves shown in Fig. \ref{fig:obstacle_loss} are reminiscent of the clutter loss model used in SAFE \cite{SAFE_PAPER}, namely the P.833 model \cite{P833}. This implies that the ML-FCN model is learning behaviours that are grounded in physics. Additionally, the expected trend of increasing loss with increasing frequency is evident in the plots.

A limitation of the ML modeling is evident by considering that obstacle loss should be insensitive to link distance (identically coloured plots should perfectly overlap in Fig. \ref{fig:obstacle_loss}) but we observe that shorter links (5 km) provide slightly higher obstacle loss than longer links (10 km). This is a limitation of the modeling that could be addressed by additional, relevant features (or additional training samples) that would help the models learn the correct influence of link distance. Another limitation of the modeling is the non-zero obstacle loss when the total obstacle depth is zero. This is likely to be in part due to LOS links still having interactions with clutter and terrain via Fresnel zone effects and NLOS paths. Both of these limitations will be addressed in future work.

%%%%%%%%%%%%% PLACEHOLDER IMAGE

\section{Conclusion}
\label{sec:conclusion}
Propagation modeling has always been integral to wireless deployments and spectrum planning with ever-increasing demands on improving modeling accuracy. Accounting for the physical environment in greater detail is key to improving prediction accuracy. In this paper, we showed that simple scalar features describing terrain and clutter can be used to train very accurate machine learning-based propagation models. When combined with link distance and frequency, simple features such as total obstacle depth along the direct path provide enough predictive ability to yield well-generalized models with RMSE in the range of 6-8 dB. Future work will address some of the limitations discussed in Section \ref{sec:obsdepthmodel} as well as incorporate clutter classification and seasonal information as inputs to the machine learning models.

% or


%\appendices

%\section{}

\section*{Acknowledgments}
The authors would like to thank Dr. Pierre Bouchard and Dr. Yvo de Jong for sharing their expertise in propagation theory and measurements, and Dr. Paul Guinand for sharing his knowledge in optimization and machine learning.

%\ifCLASSOPTIONcaptionsoff
%  \newpage
%\fi



%\bibliographystyle{IEEEtran}
%\begin{thebibliography}{1}
%\bibliography{test.bib}
%\end{thebibliography}

\bibliographystyle{IEEEtran}
\bibliography{main.bib}

%\addbibresource{test.bib}


%\end{IEEEbiography}


%\newpage



%\vfill


\end{document}